\documentclass[10pt,twocolumn,letterpaper]{article}

\usepackage{cvpr}              

\usepackage{graphicx}
\usepackage{amsmath}
\usepackage{amssymb}
\usepackage{booktabs}
\usepackage{tabularx}
\usepackage{subcaption}
\usepackage{capt-of}
\usepackage{xcolor}
\usepackage{bm}
\usepackage[accsupp]{axessibility}

\usepackage[pagebackref,breaklinks,colorlinks]{hyperref}

\newcommand{\boldstart}[1]{\noindent\textbf{#1}}
\newcommand{\suppress}[1]{}

\newcommand{\methodname}{BNV-Fusion\xspace}

\DeclareMathOperator*{\argmin}{arg\,min}

\usepackage[capitalize]{cleveref}
\crefname{section}{Sec.}{Secs.}
\Crefname{section}{Section}{Sections}
\Crefname{table}{Table}{Tables}
\crefname{table}{Tab.}{Tabs.}

\def\cvprPaperID{4177} 
\def\confName{CVPR}
\def\confYear{2022}

\begin{document}

\title{\methodname: Dense 3D Reconstruction using Bi-level Neural Volume Fusion}

\author{Kejie Li\textsuperscript{1},
 Yansong Tang\textsuperscript{1,2},
	    Victor Adrian Prisacariu\textsuperscript{1},
	    Philip H.S. Torr\textsuperscript{1}\\
\textsuperscript{1}University of Oxford,
\textsuperscript{2}Tsinghua-Berkeley Shenzhen Institute, Tsinghua University}

\twocolumn[{%
\renewcommand\twocolumn[1][]{#1}%
\maketitle
\begin{center}
\vspace{-0.7cm}
    \centering
    \includegraphics[width=0.95\linewidth]{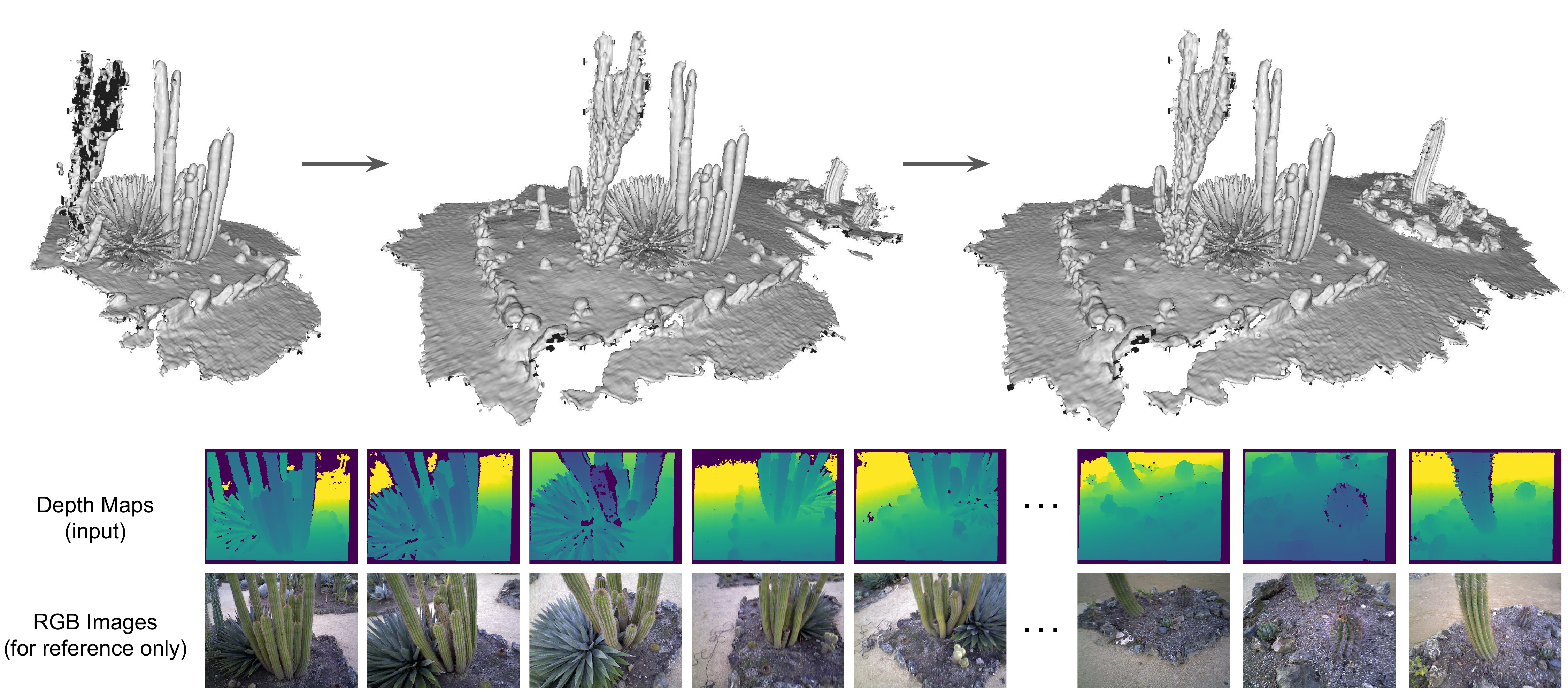}
    \captionof{figure}{\small{Bi-level Neural Volume Fusion (\methodname) incrementally integrates noisy depth images to a global model of geometry.}}
\end{center}%
}]

\begin{abstract}
Dense 3D reconstruction from a stream of depth images is the key to many mixed reality and robotic applications.
Although methods based on Truncated Signed Distance Function (TSDF) Fusion have advanced the field over the years, the TSDF volume representation is confronted with striking a balance between the robustness to noisy measurements and maintaining the level of detail.
We present Bi-level Neural Volume Fusion (\methodname), which leverages recent advances in neural implicit representations and neural rendering for dense 3D reconstruction.
In order to incrementally integrate new depth maps into a global neural implicit representation, we propose a novel bi-level fusion strategy that considers both efficiency and reconstruction quality by design. 
We evaluate the proposed method on multiple datasets quantitatively and qualitatively, demonstrating a significant improvement over existing methods.
\end{abstract}
\vspace{-0.3cm}
\section{Introduction}
Dense 3D reconstruction from images is one of the most long-standing tasks in the computer vision community. 
While there is a large body of research focusing on reconstruction using RGB-only images~\cite{Multiview,RGBsurvey}, the increasing popularity of depth sensors in commodity devices (\eg  Microsoft Kinect~\cite{MicrosoftKinect}, Apple LiDAR scanner~\cite{apple}) has enabled researchers to develop reconstruction algorithms taking advantage of depth maps~\cite{newcombe2011kinectfusion, prisacariu2017infinitam, dai2017bundlefusion}.


However, the representation used in these methods -- Truncated Signed Distance Function (TSDF) Volume -- is known to lose fine details at sub-voxel scale (\eg thin surfaces)~\cite{weder2021neuralfusion, chabra2020deep} because it discretizes the scene geometry at a pre-defined resolution.
In addition to the limitation of the representation, each depth measurement is integrated into the volume independently using voxel-wise weighted averaging without any local context, which makes the fusion process vulnerable to noisy depth measurements.

In contrast, emerging neural implicit representations, which show promising results in novel view synthesis~\cite{mildenhall2020nerf}, and shape modelling~\cite{park2019deepsdf, mescheder2019occupancy,peng2020convolutional}, have the potential of being a better alternative to TSDF volume-based reconstruction in an online setting.
In essence, these representations are deep neural networks that map continuous 3D coordinates to a task-dependent scene property, such as the color or the distance to the nearest surface.
As a result, a surface can be extracted at any resolution given the implicit function represented by the network, without any increase in memory usage.
Another advantage of neural implicit representations is that the network can be trained as a generative model to capture prior knowledge of a family of surfaces.
These appealing aspects of the neural implicit representations have motivated recent works~\cite{jiang2020local, chabra2020deep, azinovic2021neural} to develop surface reconstruction methods in an offline setting.
Nonetheless, while TSDF volume-based methods~\cite{newcombe2011kinectfusion, prisacariu2017infinitam} have demonstrated voxel-wise weighted averaging is real-time capable,
how to incrementally integrate new depth measurements using neural implicit representations is still an open question. 


Inspired by traditional volumetric fusion approaches, we present Bi-level Neural Volume Fusion (\methodname) for \textit{high-quality} and \textit{online} 3D reconstruction in this paper.
Given a sequence of depth maps and the associated poses, \methodname incrementally integrates depth measurements into a global neural volume.
The novelty of \methodname is the combination of a \textit{local}-level fusion and a \textit{global}-level fusion.
At the local level, a new depth map is first mapped to latent codes, each representing local geometry in the latent space.
They are then fused into the global neural volume by weighted averaging, which resembles the efficient update in traditional volumetric fusion methods.
However, the local-level fusion is susceptible to depth outliers as it only integrates the measured surfaces and their surroundings to the volume.
Furthermore, although several works on shape modelling~\cite{wu2016learning, girdhar2016learning} suggest that arithmetic operations in the latent space correspond to the actual geometry change to some degree, updating the global representation in the latent space using an additive scheme does not always lead to correct geometry.
To this end, we propose to optimize the global volume using neural rendering, where we penalize the discrepancies between the SDFs extracted from the global volume and those of depth measurements. 
This optimization is coined as \textit{global}-level fusion as it encourages a coherent reconstruction globally.

Overall, the key realization in \methodname is that the local- and global-level fusions are complementary.
While the local-level fusion efficiently integrates new information and initializes the global-level fusion,
the reconstruction quality is improved significantly by the global-level fusion.
To summarize, our contributions are threefold: 
\begin{itemize}
\item We propose \methodname, a novel and state-of-the-art dense 3D reconstruction pipeline that represents the geometry of a scene by an implicit neural volume. 
\item We design a novel bi-level fusion algorithm that efficiently and effectively updates the neural volume given new depth measurements.
\item We conduct extensive experiments, including an evaluation on 312 sequences of various indoor environments in ScanNet~\cite{dai2017scannet}, to validate that \methodname improves existing approaches significantly and is truly generalizable to arbitrary scenes. 
\end{itemize}

\begin{figure*}[th]
    \centering
    \includegraphics[width=0.95\linewidth]{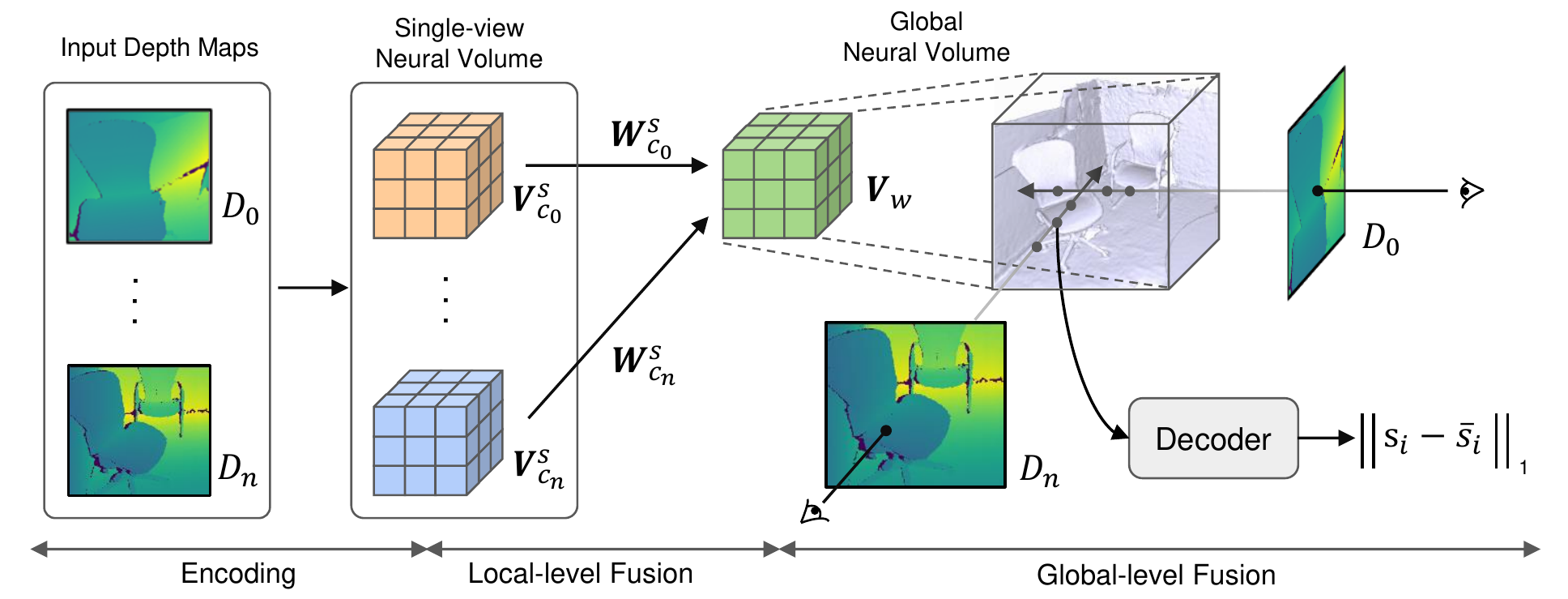}
    \caption{The architecture of \methodname. At frame $t$, \methodname first maps a depth map $D_t$ to a single-view neural volume $\bm{V}^s_{c_t}$, where each voxel contains a latent code that represents the local geometry in a latent space (\cref{sec:method_local_autoencoder}). 
    \methodname then integrates the single-view neural volume into the global neural volume using the proposed bi-level fusion (\cref{sec:method_fusion}).
    At the \textit{local} level, $\bm{V}^s_{c_t}$ is integrated into the global volume $\bm{V}_w$ using running average weighted by $\bm{W}^s_{c_t}$. At the \textit{global} level, we iteratively optimize the global volume by minimizing the discrepancies between the SDFs decoded from the volume $s_i$ and the actual measurements $\bar{s}_i$ along a camera ray.}
    \label{fig:pipeline}
\vspace{-0.2cm}
\end{figure*}
\section{Related Work}
Neural implicit representations can be categorized into global and local representations.
In ~\cref{sec:lit_rep}, we provide an overview of both categories with a focus on the latter one as the neural implicit volume in this work is inspired by methods in this category.
We then, in~\cref{sec:lit_recon}, introduce previous works in dense 3D reconstruction and describe how our method is different from existing approaches that also leverage neural implicit representations.


\subsection{Neural Implicit Representations}\label{sec:lit_rep}
\boldstart{Global representations.}
DeepSDF~\cite{park2019deepsdf}, Occupancy Networks~\cite{mescheder2019occupancy}, and IM-Net~\cite{chen2019learning} are pioneering works in neural implicit representations for object shapes.
Follow-up works~\cite{sitzmann2019scene, niemeyer2020differentiable} remove the requirement of 3D ground-truth supervision.
They train their network by minimizing the discrepancies between input images and color (and depth) images rendered from the implicit representation.
Mildenhall \etal~\cite{mildenhall2020nerf}, a seminal work in novel view synthesis using a neural implicit representation, takes a step further by representing the geometry and appearance of a scene as a Neural Radiance Field (NeRF). 
It learns to map 3D coordinates and viewing directions to occupancy and RGB values.
The idea of using neural rendering as a supervision signal to learn a neural implicit representation has inspired subsequent works in 3D reconstruction~\cite{azinovic2021neural,sucar2021imap}, including ours.

\boldstart{Local representations.}
Recent advances on implicit representations suggest that using an MLP to represent the geometry of a scene or an object is not scalable, and the prior knowledge at the object level is not generalizable~\cite{jiang2020local, chabra2020deep}.
Therefore, they propose to learn neural implicit representation for local geometry structures, which is easily generalizable to objects of novel categories. 
The geometry of an entire scene can be decomposed into a grid of local latent codes, each of which represents the geometry in the local region.
Similarly, Genova \etal~\cite{LDIF} propose a network to predict both a set of local implicit functions and their 3D locations when given a set of depth maps for object reconstruction.
In the application of novel view synthesis, Liu \etal~\cite{liu2020neural} also show that decomposing a scene into a set of local latent codes, arranged in an Octree volume, improves rendering quality and speed.
Our neural implicit volume is inspired by the work of Jiang \etal~\cite{jiang2020local} and Chabra \etal~\cite{chabra2020deep}.
While they are designed to map a complete point cloud of a scene to a set of latent codes, we can update the volume when given new information.
More importantly, their frameworks do not handle outliers explicitly as they try to fit all observed surfaces to local latent codes.
We instead filter out outliers in the proposed global-level fusion.

\subsection{Dense 3D Reconstruction}\label{sec:lit_recon}
\boldstart{Traditional approaches.}
The seminal work by Curless and Levoy~\cite{curless1996volumetric} 
presents the idea of ``TSDF Fusion'', which fuses depth maps into a TSDF volume by averaging.
KinectFusion~\cite{newcombe2011kinectfusion} revisits and extends this idea to develop a real-time dense SLAM system with commodity-level depth cameras, such as Microsoft Kinect~\cite{MicrosoftKinect}.
Subsequent works~\cite{kahler2015very,prisacariu2017infinitam,dai2017bundlefusion,Reijgwart-et-al-RAL-2020,Fuhrmann-Goesele-TOG-2011,Marniok-et-al-GCPR-2017,Steinbruecker-et-al-ICCV-2013, Niessner-et-all-ACM-2013} improve scalability, reconstruction quality, loop closure and various aspects in the fusion pipeline.
Apart from the methods based on TSDF volume, several works~\cite{Surfels, MRSMap, elasticfusion, probsurfel, SurfelMeshing} resort to surface-based representations that only model the surface of geometry for map compression.
These classic approaches have promoted the development of 3D reconstruction in the past decades. 

\boldstart{Reconstruction with neural implicit representations.}
Azinovi{\'c} \etal~\cite{azinovic2021neural} use an MLP to represent the geometry of a scene and train the MLP by comparing the rendered and input RGBD images.
Sucar \etal propose iMAP~\cite{sucar2021imap}, a dense SLAM system using a single MLP as the only representation for both mapping and tracking.
Despite shrinking the training time from days as in Azinovi{\'c} \etal~\cite{azinovic2021neural} to near real-time performance, the limitations of iMAP are as follows.
First, the reconstruction does not scale well to the size of a scene and tends to lose details.
This is because, although the MLP converges quickly to low-frequency shapes, it takes a much longer time to attend to high-frequency details, as noted by the authors of iMAP~\cite{sucar2021imap}.
Second, they train a new MLP for each scene, thereby being inefficient and prone to noise in depth measurements.
The proposed method circumvents these issues by using a volume of latent codes that encode local geometry in a shape embedding.
The volume-based representation can improve the level of detail in reconstruction because the MLP that is conditioned on a latent code only needs to learn local surface patterns rather than the geometry of a scene.
Moreover, by only optimizing the latent codes while freezing the MLP's parameters in the global-level fusion, we effectively leverage prior knowledge of local geometry embedded in MLP.

More closely related to our work as to representation are NeuralFusion~\cite{weder2021neuralfusion} and DI-Fusion~\cite{huang2021di}, both of which rely on a grid of latent codes.
However, our work differs significantly from these methods in updating the latent codes given new measurements.
Instead of only integrating new measurements in the domain of latent codes, we achieve a more globally consistent reconstruction by also optimizing the latent codes via neural rendering.

To summarize, although there are pioneering works~\cite{sucar2021imap, huang2021di, weder2021neuralfusion} that try to apply neural implicit representations to reconstruction in an online setting, they improve efficiency at a great cost of reconstruction quality.
Instead, the proposed method can reconstruct fine details that even traditional volumetric fusion approaches tend to miss while running in near real-time ($\sim 2$ Hz without proper code optimization). 
\section{Method}
Given a sequence of depth maps $\{D_0, ..., D_n\}$ and the associated extrinsic parameters $\{\bm{T}^w_{c_0}, ..., \bm{T}^w_{c_n}\}$ ($\bm{T}^w_{c_n}$ denotes a rigid transformation from camera $c_n$ to the world coordinate), \methodname aims to reconstruct the geometry of a scene represented by a global implicit neural volume $\bm{V}_w$ (defined in~\cref{sec:method_local_implicit}).
\methodname processes each depth map in three main steps, as outlined in~\cref{fig:pipeline}.
The encoding step converts a depth map into a single-view neural volume that comprises a set of latent codes in an embedding of local shapes (\cref{sec:method_local_autoencoder}).
The single-view neural volume is integrated into the global volume using the proposed bi-level fusion (\cref{sec:method_fusion}).


\subsection{Implicit Neural Volume}\label{sec:method_local_implicit}
An implicit neural volume contains a set of local latent codes spatially organized in a grid structure.
%
%
Specifically, the implicit neural volume takes the following form: $\bm{V} = \{\bm{v} = (\bm{p}, \bm{l})\}$, where each voxel $\bm{v}$ contains its 3D position in the space $\bm{p}$, and a latent code $\bm{l}$ that implicitly represents the local geometry in an embedding of local shapes.
To recover the Signed Distance Function (SDF) value of a 3D point $\bm{x}$ from the volume, we first retrieve $8$ neighboring voxels $\bm{v}_{0...7}$ in the volume and transform $\bm{x}$ to the local coordinate with respect to each neighboring voxel: $\bar{\bm{x}}_i = \bm{x} - \bm{p}_i$.
Given the latent code and the local coordinate, a shape decoder $\mathcal{D}(\cdot, \cdot)$ predicts the SDF value as follows. 
\begin{align}
    s = \sum_{i=0}^{N=7} w(\bar{\bm{x}}_i, \bm{p}_i)\mathcal{D}(\bm{l}_i, \bar{\bm{x}}_i),
\label{eq:decode_sdf}
\end{align}
where $w(\cdot, \cdot)$ is the weight of trilinear interpolation.
A mesh depicting the geometry of a scene can be extracted using the Marching Cubes algorithm~\cite{lorensen1987marching} given the SDFs decoded from the volume.

\subsection{Learning the Local Shape Embedding}\label{sec:method_local_autoencoder}
The key of the implicit neural volume to represent geometry effectively is a data-driven embedding of local shapes, which we learn using an AutoEncoder-like network.
Technically, a depth encoder $\mathcal{E}$, which is modified from PointNet~\cite{qi2017pointnet}, takes as input a point cloud unprojected from a depth map and extracts deep features of each point by aggregating information within a local region. 
These features are then mapped to a set of latent codes in the embedding.
The decoder $\mathcal{D}$ is a Multilayer Perceptron (MLP) with $4$ fully connected layers.
It takes a latent code and a 3D coordinate as input and predicts the input coordinate's SDF value.

\boldstart{Training.}
We train the encoder and decoder jointly in a supervised manner using object CAD models in ShapeNet~\cite{chang2015shapenet}.
The loss function given a pair of training sample $(\bm{p}, \bm{x})$ is 
\begin{align}
L(\theta_{\mathcal{E}}, \theta_{\mathcal{D}}) = \|\mathcal{D}_{\theta_{\mathcal{D}}}(\mathcal{E}_{\theta_{\mathcal{E}}}(\bm{p}), \bm{x}) - s^{gt}_x \|_1,
\label{eq:training_loss}
\end{align}
where $\bm{p}$ is a local point cloud with normals of a local surface patch, and $\bm{x}$ is a sampled point around the surface with its ground-truth SDF value $s^{gt}_x$.
We further detail the training process in \cref{sec:method_details}.


\begin{figure}[t]
    \centering
    \includegraphics[width=0.95\linewidth]{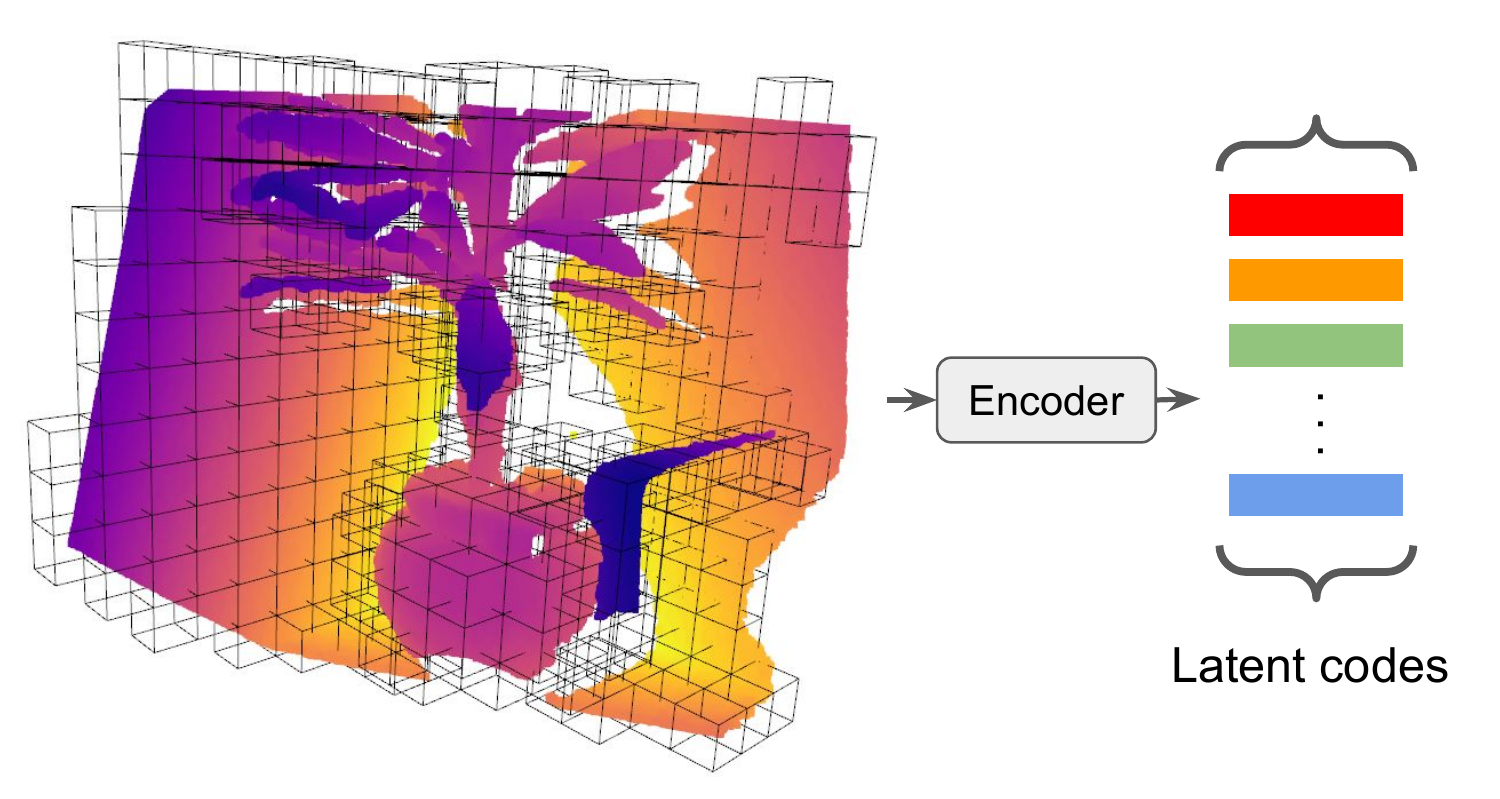}
    \caption{Encoding a depth map to the embedding of local shapes. A 3D point cloud, unprojected from a depth map, is segmented into local point clouds (bounded by 3D cuboids). The depth encoder takes as input a local point cloud and predicts a latent code.}
    \label{fig:encoding}
\vspace{-0.2cm}
\end{figure}
\boldstart{Inference.} 
At frame $t$, the depth map $D_t$ is first unprojected to a 3D point cloud using the known intrinsic parameters.
The point cloud is then segmented into overlapping  local point clouds $\{\bm{p}_0, \bm{p}_1, ..., \bm{p}_n\}$, each of which is taken by the encoder to map to a latent code $\bm{l} = \mathcal{E}(\bm{p})$, as shown in~\cref{fig:encoding}.
The latent codes are aggregated into a single-view implicit volume $\bm{V}^s_{c_t}$ that represents $D_t$ in the domain of latent space.
The implicit volume is accompanied by a weight volume $\bm{W}^s_{c_n}$, where the values are set to the number of 3D points associated with a voxel. 



\subsection{Bi-level Fusion}\label{sec:method_fusion}
Given a single-view neural volume, the global volume is updated sequentially by running the local- and global-level fusions.
The latent codes in the single-view neural volume are first integrated into the global volume by weighted averaging at the local level.
After the local update, the global volume is optimized via neural rendering to ensure a globally consistent reconstruction.
We detail the bi-level fusion in the rest of this section.

\boldstart{Local-level fusion.}
At frame $t$, the single-view neural volume $\bm{V}^s_{c_t}$ is transformed to the world coordinate using the camera extrinsic parameters: $\bm{V}^{s}_w = \bm{T}^w_{c_t} \bm{V}^{s}_{c_t}$.
The computation flow in the local-level fusion is similar to that of the traditional volumetric Fusion~\cite{curless1996volumetric} except we are averaging latent codes rather than TSDF values, as shown below:
\begin{align}
    \bm{V}^{t}_w &= \frac{\bm{W}^{t-1} \bm{V}^{t-1}_w + \bm{W}^s_{c_t} \bm{V}^{s}_w}{\bm{W}^{t-1} + \bm{W}^{s}_{c_t}}, \\
    \bm{W}^{t} &= \bm{W}^{t-1} + \bm{W}^{s}_{c_t},
\end{align}
where $\bm{V}_w$ and $\bm{W}$ denote the global implicit volume and its weight respectively. 
The superscript $t-1$ and $t$ means prior update and post update.

\boldstart{Global-level fusion.}
Although the local update is efficient, it is susceptible to outliers in measurements. 
Moreover, arithmetic operations in the latent space do not align perfectly with the actual geometry changes. 
To resolve these problems, we enforce a global consistency of geometry by rendering SDF values from the global volume and compare them with depth measurements.
The discrepancies between the rendered values and the actual observations are then used as supervision signals to optimize the global volume iteratively.
At each iteration, we randomly sample a set of pixels in a depth map. 
Along the ray unprojected from each sampled pixel, we sample $N$ 3D points $\{\bm{x}_0, \bm{x}_1, ..., \bm{x}_n\}$ using a hierarchical sampling strategy inspired by Mildenhall \etal~\cite{mildenhall2020nerf} where we sample more 3D points that are closer to the measured surface.
The projective TSDF of each 3D point given a depth measurement $\bm{p}$ on a ray is computed as: $\bar{s}_i = \min(\max(\|\bm{p} - \bm{x}_i\|_2, -\delta), \delta)$, where $\delta$ is the truncation threshold.
After transforming the 3D points to the world coordinate, the SDF values of these points can be extracted using ~\cref{eq:decode_sdf}.
Lastly, the optimization objective becomes:
\begin{align}
\argmin_{\bm{V}_w} \Sigma^{n}_{i=0} \|s_i - \bar{s}_i\|_1,
\end{align}
where $s_i$ is the SDF value decoded from the neural volume $\bm{V}_w$ using \cref{eq:decode_sdf}.


\subsection{Implementation Details}\label{sec:method_details}
\boldstart{Implicit neural volume.}
The voxel resolution is $2$ cm (\ie a latent code represents the geometry of a $2cm^{3}$ volume around its position), and the dimension of the latent codes is set to $8$.

\boldstart{Network training.}
We train the depth encoder and shape decoder using object CAD models from two categories (chairs and lamps) in ShapeNet~\cite{chang2015shapenet}.
To generate the input point clouds for the encoder, we first render $20$ depth maps from random viewpoints for each CAD model.
A depth map is unprojected to a 3D point cloud.
We then randomly select $2000$ seed points in each point cloud.
A local point cloud is created by retrieving neighboring points within a local region centered at each seed point.
We apply a Gaussian noise on the surface points and perturb their normal directions to simulate noise in depth measurements.
We also randomly sample $1000$ training points and calculate their SDF in the local region in order to train the decoder. 
The encoder contains $4$ fully connected layers with sizes [$128$, $128$, $128$, $8$], which are interconnected by ReLu except for the last layer.
We aggregate information within a local region using an average pooling layer.
The decoder is an MLP with $4$ fully connected layers with sizes [$128$, $128$, $128$, $1$].

\boldstart{Global-level fusion.}
We sample $5000$ pixels per image in each iteration. 
For each camera ray unprojected from a sampled pixel, we sample $5$ 3D points per meter at the coarse-level sampling, and $20$ 3D points at the fine-level sampling.
Latent codes in the global neural volume are optimized by the Adam optimizer~\cite{kingma2014adam} in Pytorch~\cite{paszke2019pytorch} for $5$ iterations for each image.

\begin{figure*}[th]
    \centering
    \includegraphics[width=\linewidth]{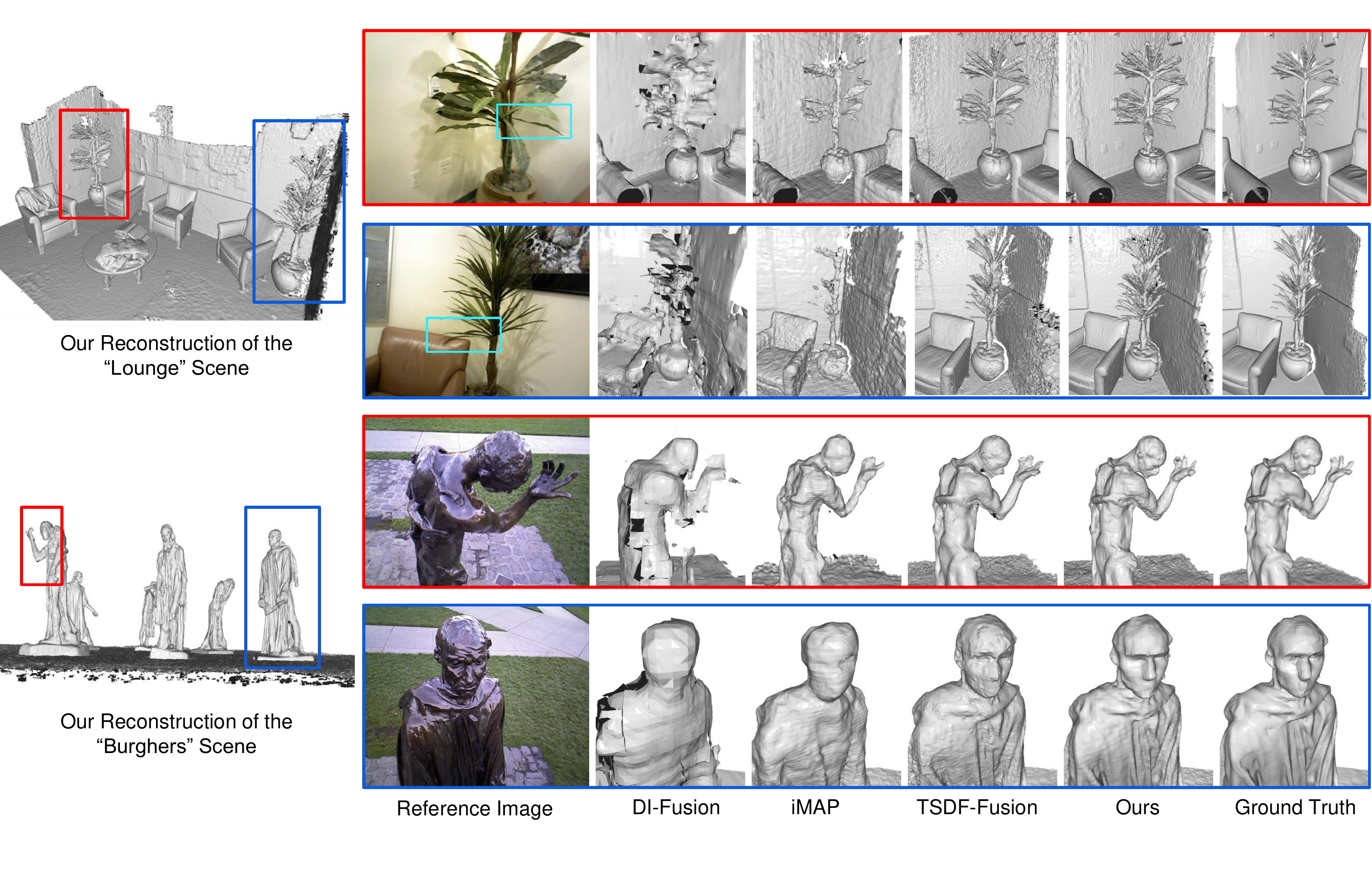}
        \vspace{-1.2cm}
    \caption{Qualitative comparison on the Scene3D dataset~\cite{Choi_2015_CVPR}. The proposed method can reconstruct more fine details accurately than previous traditional and learning-based methods. Notably, in the scene on top (the ``Lounge'' scene), our method can reconstruct some leaves and branches of the plants, which even the ground-truth mesh provided by the dataset fails to reconstruct (highlighted by the \textcolor{cyan}{cyan} bounding boxes in the reference RGB image). Compared to other methods in the scene at the bottom (the ``Burghers'' scene), our method can faithfully reconstruct the statue's face. The differences are more visible when zooming in.}
\label{fig:scene3d_qualitative}
\end{figure*}
\linespread{1.1}
\begin{table*}[t]
    \centering
    \small
    \setlength{\tabcolsep}{3.2pt}
    \begin{tabular}{l | c | c | c | c | c}
    \hline
     & Lounge & CopyRoom & CactusGarden & StoneWall & Burghers \\
    Method & Accu. / Comp. / F1 & Accu. / Comp. / F1 & Accu. / Comp. / F1 & Accu. / Comp. / F1 & Accu. / Comp. / F1 \\
    \hline
    TSDF Fusion~\cite{zhou2018open3d} & 86.16 / 93.46 / 89.66 & \textbf{89.88} / 90.22 / \textbf{90.05} & 75.62 / 94.17 / 83.84 & 88.73 / \textbf{94.34} / 91.45 & 72.78 / 82.35 / 77.26 \\
    iMAP~\cite{sucar2021imap}  & 85.76 / 87.98 / 86.85 & 83.94 / 80.22 / 82.04 & 73.04 / 85.01 / 78.57 & 85.82 / 85.83 / 85.82 & 70.23 / 71.71 / 70.96 \\
    DI-Fusion~\cite{huang2021di}  & 67.76 / 79.09 / 72.98 & 85.24 / 78.22 / 81.58 & 58.00 / 68.70 / 62.90 & 82.36 / 89.97 / 85.90 & 63.10 / 65.90 / 64.47 \\
    \methodname (Ours)  & \textbf{87.53} / \textbf{94.77} / \textbf{91.01} & 88.56 / \textbf{90.32} / 89.43 & \textbf{78.62} / \textbf{94.33} / \textbf{85.75} & \textbf{92.57} / 94.19 / \textbf{93.37} & \textbf{75.98} / \textbf{82.44} / \textbf{79.08} \\
    \hline
    \end{tabular}
    \caption{Quantitative evaluation on the 3D Scene Dataset~\cite{Choi_2015_CVPR}. Our BNVF shows superior performance over the state-of-the-arts.}
    \label{table:3d_scene}
\vspace{-0.3cm}
\end{table*}
\linespread{1}

\section{Experiments}
\subsection{Datasets and Metrics}
We evaluate the proposed method extensively on three datasets: 3D Scene Dataset~\cite{Choi_2015_CVPR}, ICL-NUIM RGBD benchmark~\cite{handa_ICRA2014} (under the Creative Commons 3.0 license), and ScanNet~\cite{dai2017scannet} (under the MIT license). 
These datasets cover both synthetic scans with ground-truth 3D models, and real-world scans with pseudo ground-truth 3D models.
To evaluate reconstruction quality, we uniformly sample $100,000$ points from the ground-truth meshes and reconstructed meshes, respectively, then report the following metrics. 
\textit{Accuracy} (denoted as Accu. in tables) measures the fraction of points from the reconstructed mesh that are closer to points from the ground-truth mesh than a threshold distance, which is set to $2.5$ cm.
\textit{Completeness} (denoted as Comp. in tables) calculates the fraction of points from the ground-truth mesh that are closer to points from the reconstructed mesh than $2.5$ cm.
\textit{F1 score} (denoted as F1 in tables) is the harmonic mean of accuracy and completeness, which quantifies the overall reconstruction quality.
When reporting per-sequence quantitative results on ICL-NUIM~\cite{handa_ICRA2014} and the 3D Scene dataset~\cite{Choi_2015_CVPR}, we run experiments $5$ times using different frames and sample surface points twice independently to ensure the quantitative results are statistically significant.

\subsection{Baselines}
In the following experiments, we demonstrate the effectiveness of \methodname by comparing against three strong baseline methods that use traditional TSDF fusion or build upon modern neural implicit representations.

``TSDF Fusion'' in this section denotes an implementation of TSDF Fusion in the Open3D library~\cite{zhou2018open3d} based on KinectFusion~\cite{newcombe2011kinectfusion}.
``DI-Fusion''~\cite{huang2021di} is a reconstruction pipeline using a volumetric neural implicit representation.
We use the code published by the authors in our experiments.
``iMAP'' is our reimplementation of the paper iMAP from Sucar \etal~\cite{sucar2021imap} because we do not have access to the official implementation at the time of submission. 
Running our system on their dataset is not preferable since they use synthetic depth maps without noise in evaluation, which have a domain gap to real-world scans.
To isolate the mapping components in all compared methods, we use camera poses provided by the datasets and take the same images as input in the evaluation.

\begin{table*}[t]
    \centering
    \small
    \setlength{\tabcolsep}{4.5mm}{
    \begin{tabular}{l | c | c | c | c}
    \hline
     & Livingroom0 & Livingroom1 & Office0 & Office1 \\
    Method & Accu. / Comp. / F1 & Accu. / Comp. / F1 & Accu. / Comp. / F1 & Accu. / Comp. / F1e \\
    \hline
    TSDF Fusion~\cite{zhou2018open3d} & 54.66 / 62.52 / 58.31 & 60.13 / 72.19 / 65.28 & 49.21 / 53.71 / 51.41 & 56.61 / 59.18 / 57.79 \\
    iMAP~\cite{sucar2021imap} & 61.23 / 62.31 / 61.76 & 65.12 / 65.40 / 62.25 & 47.26 / 47.18 / 47.22 & \textbf{64.96} / 59.91 / 57.33 \\
    DI-Fusion~\cite{huang2021di} & 61.52 / 64.43 / 62.55 & 69.69 / 67.81 / 68.00 & 50.06 / 48.99 / 49.79 & 54.53 / 50.41 / 52.36 \\
    BNVF (Ours) & \textbf{71.26} / \textbf{73.86} / \textbf{72.54} & \textbf{80.01} / \textbf{81.94} / \textbf{81.02} & \textbf{58.17} / \textbf{60.33} / \textbf{59.23} & 63.93 / \textbf{63.89} / \textbf{63.91} \\ 
    \hline
    \end{tabular}
    }
    \caption{Comparison of accuracy (Accu.), completeness (Comp.) and F1 score (F1) on the augmented ICL-NUIM dataset~\cite{Choi_2015_CVPR}. The proposed \methodname outperforms all methods by a large margin in all metrics in $4$ sequences except the accuracy in the ``Office1'' sequence.}
    \label{tab:icl_nuim}
\vspace{-0.2cm}
\end{table*}
\linespread{1}
\begin{table}[t]
    \centering
    \scriptsize
    \setlength{\tabcolsep}{3.0pt}
    \begin{tabular}{l||c|c||c|c||c}
        Methods & \multicolumn{2}{c||}{Thresholds} & \multicolumn{2}{c||}{Every $x^{th}$ frames @ $2.5$cm} & Estimated Poses \\
        \hline
         & $1$cm & $5$cm & $x=1$ & $x=30$ \\ 
        \hline
        TSDF~\cite{zhou2018open3d} & 33.63 & 90.87 & 85.26 & 79.81 & 55.20 \\
        iMAP~\cite{sucar2021imap} & 28.43 & 89.72 & 82.07 & 74.12 & 55.96 \\
        DI-Fusion~\cite{huang2021di} & 23.53 & 88.83 & 74.40 & 73.81 & 58.33 \\
        Ours & \textbf{35.34} & \textbf{93.29} & \textbf{87.60} & \textbf{80.57} & \textbf{61.07}
    \vspace{-0.2cm}
    \end{tabular}
    \caption{F1 scores under various experimental settings.}
    \label{tab:extra}
\end{table}

\subsection{Evaluation on 3D Scene Dataset}
The 3D Scene dataset, which comprises several real-world RGBD sequences, is a popular benchmark in the reconstruction community. 
Unlike synthetic datasets (\eg the ICL-NUIM dataset), where ground-truth 3D models are available, in order to provide a quantitative comparison, we follow a common practice in prior art~\cite{chabra2020deep, weder2021neuralfusion}.
Specifically, we consider the reconstructions provided by the dataset as ground truth.
Since those models are created by running a TSDF Fusion method on all available frames in each sequence with post-processing, we take only every $10^{th}$ frame as input images when running compared methods in the evaluation.

~\cref{table:3d_scene} quantifies the reconstruction quality on $5$ scenes in the 3D Scene dataset, from which it is clear that our method outperforms other methods in all sequences.
We highlight the differences in reconstructions produced by different methods in ~\cref{fig:scene3d_qualitative}.
Although DI-Fusion~\cite{huang2021di} is able to reconstruct smooth surfaces given noisy depth measurements, it fails to capture any fine details.
This suggests that integrating depth measurements in the latent space only does not utilize the depth measurements effectively.
While decreasing the voxel size in DI-Fusion seems to be an alternative to facilitate accurate reconstruction, we present an ablation study on voxel size in~\cref{sec:exp_ablation} to show that there is a problematic trade-off if reducing voxel size.
iMAP~\cite{sucar2021imap}, another method based on a neural implicit representation, is unable to attend to details either because they use a single MLP to represent the entire scene.
It is also visible in ~\cref{fig:scene3d_qualitative} that iMAP is more susceptible to noise in depth maps than both our method and DI-Fusion.
This is due to the lack of prior knowledge since they train a new network for each scene. 
Compared to the TSDF-Fusion approach, our method reconstructs challenging structures more accurately, such as thin leaves, fingertips, and human faces.
Furthermore, by referring to the RGB images, it is encouraging to see that our method can even reconstruct thin structures (highlighted in the cyan bounding boxes in~\cref{fig:scene3d_qualitative}) that the ground-truth meshes miss.
Note that $10\times$ more depth images are used to generate the ground-truth meshes.
In addition, \cref{tab:extra} evaluates the methods under different experiment settings (\eg using SLAM to track camera, different thresholds, different frame rates).


\subsection{Evaluation on ICL-NUIM Dataset}
The ICL-NUIM dataset is a synthetic dataset with ground-truth 3D models. 
We use the synthetic sequences rendered by Choi \etal~\cite{Choi_2015_CVPR} because their rendering considers a more comprehensive noise model 
(\eg disparity-based quantization, realistic high-frequency noise, and low-frequency distortion based on real depth cameras) 
to simulate the noise in real depth images.
Our reconstructions are more accurate and complete, supported by the high accuracy and completeness in ~\cref{tab:icl_nuim}.
Similar to the results in the 3D Scene Dataset, both DI-Fusion~\cite{huang2021di} and iMAP~\cite{sucar2021imap} struggle to reconstruct fine details in the geometry.
We outperform TSDF-Fusion~\cite{zhou2018open3d} by a larger margin than what we have in the 3D Scene Dataset.
We believe this is because the synthetic noise in ICL-NUIM dataset is higher than the actual noise in a real depth sensor for large depth values, which indicates that our method is robust to greater noise in measurements.
We present qualitative results of all compared methods in the supplementary material.

\subsection{Evaluation on ScanNet}

\linespread{1.1}
\begin{table}[t]
    \centering
    \small
    \setlength{\tabcolsep}{4mm}{
    \begin{tabular}{l | c c c }
    \hline
    Method &  Accu. & Comp. & F1 \\
    \hline
    TSDF Fusion~\cite{zhou2018open3d} & 73.83 & 85.85 & 78.84 \\
    iMAP & 68.96 & 82.12 & 74.96 \\
    DI-Fusion & 66.34 & 79.65 & 72.97 \\
    BNVF (Ours) & \textbf{74.90} & \textbf{88.12} & \textbf{80.56} \\
    \hline
    \end{tabular}
    }
    \caption{Comparison between our proposed \methodname and TSDF-Fusion on ScanNet~\cite{dai2017scannet}. It demonstrates that the proposed method can generalize well to various scenes.}
    \label{tab:scannet}
\vspace{-0.2cm}
\end{table}
\linespread{1}

Both the ICL-NUIM and 3D Scene datasets have only a handful of sequences, which might not be diverse enough to test the generalization of a learning-based method.
To this end, we evaluate our method on the validation set of  ScanNet~\cite{dai2017scannet}, comprising 312 sequences captured in diverse indoor environments, such as living rooms, conference rooms, and offices.
Since ScanNet also does not have ground-truth models of the scenes, we again use every $10^{th}$ frame as input and consider the meshes provided by the dataset as pseudo ground truth. 
The quantitative comparison is summarized in~\cref{tab:scannet}.
Outperforming TSDF Fusion in various scenes in ScanNet demonstrates the excellent generalization of the proposed method.
We present a few example reconstructions in the supplementary material.

\subsection{Ablation studies}\label{sec:exp_ablation}
The proposed method differs from previous online reconstruction frameworks that use neural implicit representations thanks to the combination of the local- and global-level fusion. 
Therefore, we present more analysis on the bi-level fusion in this section using sequences in the augmented ICL-NUIM dataset~\cite{Choi_2015_CVPR}.

\linespread{1.1}
\begin{table}[t]
    \centering
    \small
    \setlength{\tabcolsep}{3.5mm}{
    \begin{tabular}{l | c c c}
    \hline
    Method &  Accu. & Comp. & F1 \\
    \hline
    w/o global-level fusion & 30.29 & 39.64 & 34.31 \\
    w/o local-level fusion & 33.61 & 38.50 & 36.55 \\
    Ours (local + global) & \textbf{68.34} & \textbf{70.01} & \textbf{69.17} \\
    \hline
    \end{tabular}
    }
    \caption{A quantitative comparison of different fusion algorithms. The bi-level fusion significantly outperforms both baselines.}
    \label{tab:ablation}
\vspace{-0.3cm}
\end{table}
\linespread{1}


\begin{figure*}
  \centering
  \includegraphics[width=0.95\linewidth]{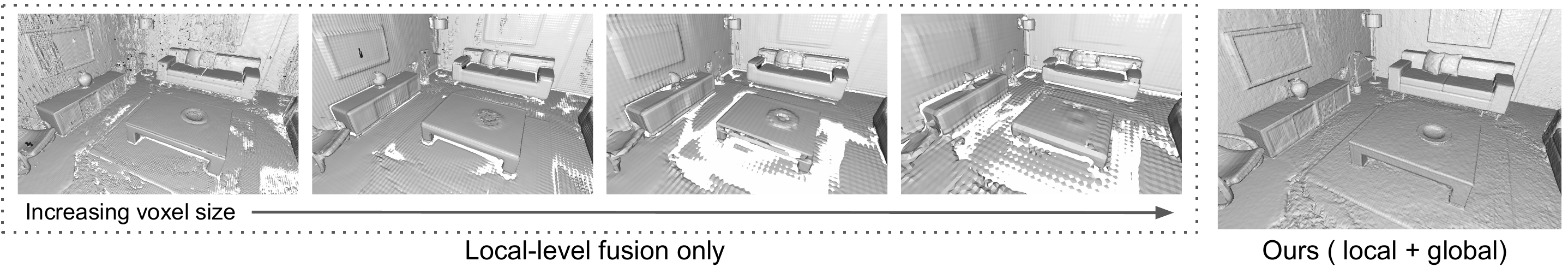}
  \caption{A visual comparison between local-level fusion only and the proposed method. Reconstructions using local-level fusion only clearly suffer from outliers and missing structures. Increasing the size of implicit voxels, from $5$ cm to $20$ cm with a step size $5$ cm, as shown in the figure, reduces outliers at the cost of losing even more details.}
 \label{fig:local_level_ablation}
\vspace{-0.15cm}
\end{figure*}
\boldstart{Global-level fusion.}
Compared to prior art~\cite{weder2021neuralfusion, huang2021di} that only fuses information in the latent space, we optimize the global volume using neural rendering to achieve consistent reconstruction.
In addition to the comparison against prior art reported in~\cref{table:3d_scene} and~\cref{tab:icl_nuim}, we validate the necessity of the global-level fusion by disabling it in our system. 

The quantitative comparison is reported in ~\cref{tab:ablation}, and the contrast is visualized in ~\cref{fig:local_level_ablation}.
Using local-level fusion only with small voxel size suffers from depth outliers because it updates each implicit voxel independently in the domain of latent codes.
We also test an approach used in DI-Fusion~\cite{huang2021di}, which is increasing the voxel size.
However, as shown in ~\cref{fig:local_level_ablation}, this is also undesirable since more details are lost as the voxel size increases.

\boldstart{Local-level fusion.}
\begin{figure}[t]
\centering
\includegraphics[width=0.95\linewidth]{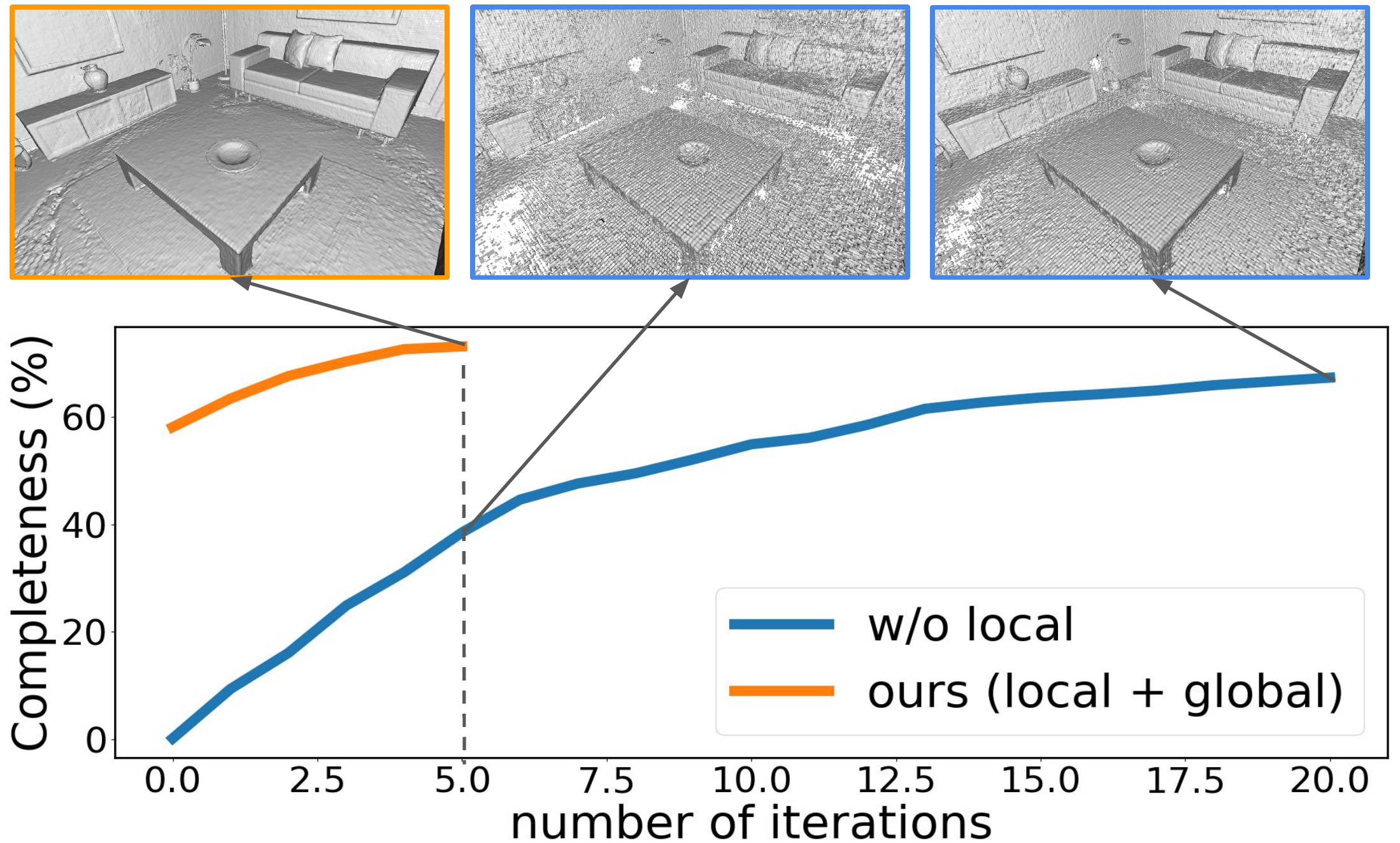}
\caption{Improvement in completeness as the global-level fusion progresses. The global-level fusion initialized by the local-level fusion (shown in \textcolor{orange}{orange} line) converges in $5$ iterations. In contrast, the global-level fusion initialized randomly (shown in \textcolor{blue}{blue} line) takes a much longer time to converge, and it is still not as good as the proposed method even after $20$ iterations.}
\label{fig:ablate_local}
\vspace{-0.2cm}
\end{figure}
One of the key contributions of the local-level fusion is to initialize the global-level fusion.
We compare our method against a baseline that initializes the global-level fusion randomly.
~\cref{fig:ablate_local} illustrates that the baseline requires more iterations to converge that our method.
We also quantitatively compare the reconstructions of the baseline and that of our method by running the same number of iterations in ~\cref{tab:ablation}, which clearly shows that the local-level fusion is crucial for dense reconstruction in an online setting.




\subsection{Runtime analysis}
We break down the runtime of each component as follows. 
Encoding a depth map to the latent space and the local-level fusion takes $0.1$ seconds in total, the majority of which is taken by the encoding step.
Running the global-level fusion for $2$ iterations takes $0.5$ seconds.
Mesh extraction is excluded since it can be run on a separate thread.
Overall, \methodname runs at almost $2$ frames per second (fps)  on a 1080Ti GPU.
DI-Fusion~\cite{huang2021di} runs at $~10$ fps on the same device.
iMAP~\cite{sucar2021imap} is not real-time capable because we run on all frames rather than selected keyframes in the original paper.

\subsection{Limitations}\label{sec:limitation}
There are two limitations to be considered.
First, the proposed method is still slower than traditional volumetric fusion approaches, which are heavily engineered.
For instance, InfiniTAM v3~\cite{InfiniTAM_ISMAR_2015} can easily run at over $30$ fps on a mobile device.
Future research will need to be done to develop a neural-implicit-based reconstruction pipeline that can be run as fast as the traditional approaches.
Second, severe noise in depth measurements sometimes causes discontinuities between neighboring implicit voxels, even though we try to improve the border consistency by using trilinear interpolation among neighboring implicit voxels.

\section{Conclusion}
We present \methodname, a novel online approach that effectively uses implicit neural volumes to represent geometry, for 3D reconstruction.
The core of \methodname is the bi-level fusion algorithm:
1) The local-level fusion efficiently fuses new depth maps into the global volume;
2) The global-level fusion, framed as neural rendering, facilitates a consistent reconstruction.
We evaluate \methodname in multiple reconstruction benchmarks, where it shows significant improvements in accuracy and completion over both a traditional volumetric fusion approach and recent learning-based approaches. 
This verifies \methodname's capability of reconstructing the geometry faithfully.
Furthermore, we justify the bi-level fusion in the ablation studies.
Despite limitations discussed in~\cref{sec:limitation}, the superior performance of \methodname is still encouraging, hence we believe this work shows notable progress in dense 3D reconstruction.

\boldstart{Acknowledgement}
This work is supported by the UKRI grant: Turing AI Fellowship EP/W002981/1 and EPSRC/MURI grant: EP/N019474/1. We would also like to thank the Royal Academy of Engineering and FiveAI.
\clearpage

{\small
\bibliographystyle{ieee_fullname}
\bibliography{reference}

\begin{thebibliography}{10}\itemsep=-1pt

\bibitem{apple}
Apple {L}i{DAR} scanner.
\newblock
  https://www.apple.com/newsroom/\\2020/03/apple-unveils-new-ipad-pro-with-lidar-scanner-and-trackpad-support-in-ipados.

\bibitem{azinovic2021neural}
Dejan Azinovi{\'c}, Ricardo Martin-Brualla, Dan~B Goldman, Matthias
  Nie{\ss}ner, and Justus Thies.
\newblock Neural rgb-d surface reconstruction.
\newblock {\em arXiv preprint arXiv:2104.04532}, 2021.

\bibitem{chabra2020deep}
Rohan Chabra, Jan~E Lenssen, Eddy Ilg, Tanner Schmidt, Julian Straub, Steven
  Lovegrove, and Richard Newcombe.
\newblock Deep local shapes: Learning local sdf priors for detailed 3d
  reconstruction.
\newblock In {\em European Conference on Computer Vision}, pages 608--625.
  Springer, 2020.

\bibitem{chang2015shapenet}
Angel~X Chang, Thomas Funkhouser, Leonidas Guibas, Pat Hanrahan, Qixing Huang,
  Zimo Li, Silvio Savarese, Manolis Savva, Shuran Song, Hao Su, et~al.
\newblock Shapenet: An information-rich 3d model repository.
\newblock {\em arXiv preprint arXiv:1512.03012}, 2015.

\bibitem{chen2019learning}
Zhiqin Chen and Hao Zhang.
\newblock Learning implicit fields for generative shape modeling.
\newblock In {\em Proceedings of the IEEE/CVF Conference on Computer Vision and
  Pattern Recognition}, pages 5939--5948, 2019.

\bibitem{Choi_2015_CVPR}
Sungjoon Choi, Qian-Yi Zhou, and Vladlen Koltun.
\newblock Robust reconstruction of indoor scenes.
\newblock In {\em Proceedings of the IEEE/CVF Conference on Computer Vision and
  Pattern Recognition}, 2015.

\bibitem{curless1996volumetric}
Brian Curless and Marc Levoy.
\newblock A volumetric method for building complex models from range images.
\newblock In {\em Proceedings of the 23rd annual conference on Computer
  graphics and interactive techniques}, pages 303--312, 1996.

\bibitem{dai2017scannet}
Angela Dai, Angel~X Chang, Manolis Savva, Maciej Halber, Thomas Funkhouser, and
  Matthias Nie{\ss}ner.
\newblock Scannet: Richly-annotated 3d reconstructions of indoor scenes.
\newblock In {\em Proceedings of the IEEE conference on Computer Vision and
  Pattern Recognition}, pages 5828--5839, 2017.

\bibitem{dai2017bundlefusion}
Angela Dai, Matthias Nie{\ss}ner, Michael Zollh{\"o}fer, Shahram Izadi, and
  Christian Theobalt.
\newblock Bundlefusion: Real-time globally consistent 3d reconstruction using
  on-the-fly surface reintegration.
\newblock {\em ACM Transactions on Graphics (ToG)}, 36(4):1, 2017.

\bibitem{Fuhrmann-Goesele-TOG-2011}
Simon Fuhrmann and Michael Goesele.
\newblock Fusion of depth maps with multiple scales.
\newblock {\em {ACM} Trans. Graph.}, 30(6):148:1--148:8, 2011.

\bibitem{LDIF}
Kyle Genova, Forrester Cole, Avneesh Sud, Aaron Sarna, and Thomas~A.
  Funkhouser.
\newblock Local deep implicit functions for 3d shape.
\newblock In {\em Proceedings of the IEEE/CVF Conference on Computer Vision and
  Pattern Recognition}, pages 4856--4865, 2020.

\bibitem{girdhar2016learning}
Rohit Girdhar, David~F Fouhey, Mikel Rodriguez, and Abhinav Gupta.
\newblock Learning a predictable and generative vector representation for
  objects.
\newblock In {\em European Conference on Computer Vision}, pages 484--499.
  Springer, 2016.

\bibitem{RGBsurvey}
Xian{-}Feng Han, Hamid Laga, and Mohammed Bennamoun.
\newblock Image-based 3d object reconstruction: State-of-the-art and trends in
  the deep learning era.
\newblock {\em {IEEE} Trans. Pattern Anal. Mach. Intell.}, 43(5):1578--1604,
  2021.

\bibitem{handa_ICRA2014}
A. Handa, T. Whelan, J.B. McDonald, and A.J. Davison.
\newblock A benchmark for {RGB-D} visual odometry, {3D} reconstruction and
  {SLAM}.
\newblock In {\em IEEE Intl. Conf. on Robotics and Automation, ICRA}, Hong
  Kong, China, May 2014.

\bibitem{Multiview}
Andrew Harltey and Andrew Zisserman.
\newblock {\em Multiple view geometry in computer vision}.
\newblock Cambridge University Press, 2003.

\bibitem{huang2021di}
Jiahui Huang, Shi-Sheng Huang, Haoxuan Song, and Shi-Min Hu.
\newblock Di-fusion: Online implicit 3d reconstruction with deep priors.
\newblock In {\em Proceedings of the IEEE/CVF Conference on Computer Vision and
  Pattern Recognition}, pages 8932--8941, 2021.

\bibitem{jiang2020local}
Chiyu Jiang, Avneesh Sud, Ameesh Makadia, Jingwei Huang, Matthias Nie{\ss}ner,
  Thomas Funkhouser, et~al.
\newblock Local implicit grid representations for 3d scenes.
\newblock In {\em Proceedings of the IEEE/CVF Conference on Computer Vision and
  Pattern Recognition}, pages 6001--6010, 2020.

\bibitem{kahler2015very}
Olaf K{\"a}hler, Victor~Adrian Prisacariu, Carl~Yuheng Ren, Xin Sun, Philip
  Torr, and David Murray.
\newblock Very high frame rate volumetric integration of depth images on mobile
  devices.
\newblock {\em IEEE transactions on visualization and computer graphics},
  21(11):1241--1250, 2015.

\bibitem{InfiniTAM_ISMAR_2015}
O. Kahler, V.~A. Prisacariu, C.~Y. Ren, X. Sun, P.~H.~S Torr, and D.~W. Murray.
\newblock {Very High Frame Rate Volumetric Integration of Depth Images on
  Mobile Device}.
\newblock {\em IEEE Transactions on Visualization and Computer Graphics},
  22(11), 2015.

\bibitem{kingma2014adam}
Diederik~P Kingma and Jimmy Ba.
\newblock Adam: A method for stochastic optimization.
\newblock {\em Proceedings of the International Conference on Learning
  Representations (ICLR)}, 2015.

\bibitem{liu2020neural}
Lingjie Liu, Jiatao Gu, Kyaw~Zaw Lin, Tat-Seng Chua, and Christian Theobalt.
\newblock Neural sparse voxel fields.
\newblock {\em arXiv preprint arXiv:2007.11571}, 2020.

\bibitem{lorensen1987marching}
William~E Lorensen and Harvey~E Cline.
\newblock Marching cubes: A high resolution 3d surface construction algorithm.
\newblock {\em ACM siggraph computer graphics}, 21(4):163--169, 1987.

\bibitem{Marniok-et-al-GCPR-2017}
Nico Marniok, Ole Johannsen, and Bastian Goldluecke.
\newblock An efficient octree design for local variational range image fusion.
\newblock pages 401--412, 2017.

\bibitem{mescheder2019occupancy}
Lars Mescheder, Michael Oechsle, Michael Niemeyer, Sebastian Nowozin, and
  Andreas Geiger.
\newblock Occupancy networks: Learning 3d reconstruction in function space.
\newblock In {\em Proceedings of the IEEE/CVF Conference on Computer Vision and
  Pattern Recognition}, pages 4460--4470, 2019.

\bibitem{mildenhall2020nerf}
Ben Mildenhall, Pratul~P Srinivasan, Matthew Tancik, Jonathan~T Barron, Ravi
  Ramamoorthi, and Ren Ng.
\newblock Nerf: Representing scenes as neural radiance fields for view
  synthesis.
\newblock In {\em European conference on computer vision}, pages 405--421.
  Springer, 2020.

\bibitem{newcombe2011kinectfusion}
Richard~A Newcombe, Shahram Izadi, Otmar Hilliges, David Molyneaux, David Kim,
  Andrew~J Davison, Pushmeet Kohi, Jamie Shotton, Steve Hodges, and Andrew
  Fitzgibbon.
\newblock Kinectfusion: Real-time dense surface mapping and tracking.
\newblock In {\em 2011 10th IEEE international symposium on mixed and augmented
  reality}, pages 127--136. IEEE, 2011.

\bibitem{niemeyer2020differentiable}
Michael Niemeyer, Lars Mescheder, Michael Oechsle, and Andreas Geiger.
\newblock Differentiable volumetric rendering: Learning implicit 3d
  representations without 3d supervision.
\newblock In {\em Proceedings of the IEEE/CVF Conference on Computer Vision and
  Pattern Recognition}, pages 3504--3515, 2020.

\bibitem{Niessner-et-all-ACM-2013}
Matthias Nie{\ss}ner, Michael Zollh{\"{o}}fer, Shahram Izadi, and Marc
  Stamminger.
\newblock Real-time 3d reconstruction at scale using voxel hashing.
\newblock {\em {ACM} Trans. Graph.}, 32(6):169:1--169:11, 2013.

\bibitem{park2019deepsdf}
Jeong~Joon Park, Peter Florence, Julian Straub, Richard Newcombe, and Steven
  Lovegrove.
\newblock Deepsdf: Learning continuous signed distance functions for shape
  representation.
\newblock In {\em Proceedings of the IEEE/CVF Conference on Computer Vision and
  Pattern Recognition}, pages 165--174, 2019.

\bibitem{paszke2019pytorch}
Adam Paszke, Sam Gross, Francisco Massa, Adam Lerer, James Bradbury, Gregory
  Chanan, Trevor Killeen, Zeming Lin, Natalia Gimelshein, Luca Antiga, et~al.
\newblock Pytorch: An imperative style, high-performance deep learning library.
\newblock {\em Advances in neural information processing systems},
  32:8026--8037, 2019.

\bibitem{peng2020convolutional}
Songyou Peng, Michael Niemeyer, Lars Mescheder, Marc Pollefeys, and Andreas
  Geiger.
\newblock Convolutional occupancy networks.
\newblock In {\em European Conference on Computer Vision}, pages 523--540.
  Springer, 2020.

\bibitem{Surfels}
Hanspeter Pfister, Matthias Zwicker, Jeroen van Baar, and Markus~H. Gross.
\newblock Surfels: surface elements as rendering primitives.
\newblock In Judith~R. Brown and Kurt Akeley, editors, {\em SIGGRAPH}, pages
  335--342.

\bibitem{prisacariu2017infinitam}
Victor~Adrian Prisacariu, Olaf K{\"a}hler, Stuart Golodetz, Michael Sapienza,
  Tommaso Cavallari, Philip~HS Torr, and David~W Murray.
\newblock Infinitam v3: A framework for large-scale 3d reconstruction with loop
  closure.
\newblock {\em arXiv preprint arXiv:1708.00783}, 2017.

\bibitem{qi2017pointnet}
Charles~R Qi, Hao Su, Kaichun Mo, and Leonidas~J Guibas.
\newblock Pointnet: Deep learning on point sets for 3d classification and
  segmentation.
\newblock In {\em Proceedings of the IEEE conference on computer vision and
  pattern recognition}, pages 652--660, 2017.

\bibitem{Reijgwart-et-al-RAL-2020}
V. {Reijgwart}, A. {Millane}, H. {Oleynikova}, R. {Siegwart}, C. {Cadena}, and
  J. {Nieto}.
\newblock Voxgraph: Globally consistent, volumetric mapping using signed
  distance function submaps.
\newblock {\em IEEE Robotics and Automation Letters}, 2020.

\bibitem{SurfelMeshing}
Thomas Sch{\"{o}}ps, Torsten Sattler, and Marc Pollefeys.
\newblock Surfelmeshing: Online surfel-based mesh reconstruction.
\newblock {\em {IEEE} Trans. Pattern Anal. Mach. Intell.}, 42(10):2494--2507,
  2020.

\bibitem{sitzmann2019scene}
Vincent Sitzmann, Michael Zollh{\"o}fer, and Gordon Wetzstein.
\newblock Scene representation networks: Continuous 3d-structure-aware neural
  scene representations.
\newblock {\em arXiv preprint arXiv:1906.01618}, 2019.

\bibitem{Steinbruecker-et-al-ICCV-2013}
Frank Steinbr{\"{u}}cker, Christian Kerl, and Daniel Cremers.
\newblock Large-scale multi-resolution surface reconstruction from {RGB-D}
  sequences.
\newblock In {\em Proceedings of the IEEE/CVF International Conference on
  Computer Vision}, pages 3264--3271, 2013.

\bibitem{MRSMap}
J{\"{o}}rg St{\"{u}}ckler and Sven Behnke.
\newblock Multi-resolution surfel maps for efficient dense 3d modeling and
  tracking.
\newblock {\em J. Vis. Commun. Image Represent.}, 25(1):137--147, 2014.

\bibitem{sucar2021imap}
Edgar Sucar, Shikun Liu, Joseph Ortiz, and Andrew~J Davison.
\newblock imap: Implicit mapping and positioning in real-time.
\newblock In {\em Proceedings of the IEEE/CVF International Conference on
  Computer Vision}, pages 6229--6238, 2021.

\bibitem{weder2021neuralfusion}
Silvan Weder, Johannes~L Schonberger, Marc Pollefeys, and Martin~R Oswald.
\newblock Neuralfusion: Online depth fusion in latent space.
\newblock In {\em Proceedings of the IEEE/CVF Conference on Computer Vision and
  Pattern Recognition}, pages 3162--3172, 2021.

\bibitem{elasticfusion}
Thomas Whelan, Renato~F. Salas{-}Moreno, Ben Glocker, Andrew~J. Davison, and
  Stefan Leutenegger.
\newblock Elasticfusion: Real-time dense {SLAM} and light source estimation.
\newblock {\em Int. J. Robotics Res.}, 35(14):1697--1716, 2016.

\bibitem{wu2016learning}
Jiajun Wu, Chengkai Zhang, Tianfan Xue, William~T Freeman, and Joshua~B
  Tenenbaum.
\newblock Learning a probabilistic latent space of object shapes via 3d
  generative-adversarial modeling.
\newblock In {\em Proceedings of the 30th International Conference on Neural
  Information Processing Systems}, pages 82--90, 2016.

\bibitem{probsurfel}
Zhixin Yan, Mao Ye, and Liu Ren.
\newblock Dense visual {SLAM} with probabilistic surfel map.
\newblock {\em {IEEE} Trans. Vis. Comput. Graph.}, 23(11):2389--2398, 2017.

\bibitem{MicrosoftKinect}
Zhengyou Zhang.
\newblock Microsoft kinect sensor and its effect.
\newblock {\em {IEEE} Multim.}, 19(2):4--10, 2012.

\bibitem{zhou2018open3d}
Qian-Yi Zhou, Jaesik Park, and Vladlen Koltun.
\newblock Open3d: A modern library for 3d data processing.
\newblock {\em arXiv preprint arXiv:1801.09847}, 2018.

\end{thebibliography}
}

\end{document}